\documentclass[letterpaper, 10 pt, conference]{ieeeconf}  

\IEEEoverridecommandlockouts                              

\usepackage{amsmath} 
\usepackage{amssymb}  
\usepackage{graphicx}
\usepackage[linesnumbered,ruled,vlined]{algorithm2e}
\usepackage{caption}
\usepackage{hyperref}
\let\labelindent\relax
\usepackage{enumitem}
\usepackage{subcaption}
\usepackage{multirow}
\usepackage{bm}
\usepackage{mathtools}

\newcommand{\link}[1]{{\href{#1}{#1}}}

\title{\LARGE \bf
ExploRLLM: Guiding Exploration in Reinforcement Learning with Large Language Models
}

\author{
    Runyu Ma\textsuperscript{*1}, Jelle Luijkx\textsuperscript{*1}, Zlatan Ajanovi{\'c}\textsuperscript{2}, and Jens Kober\textsuperscript{1}
    \thanks{
    \textsuperscript{*} Equal Contribution.
    \textsuperscript{1} Cognitive Robotics, Delft University of Technology, The Netherlands (e-mail: \{j.d.luijkx, j.kober\}@tudelft.nl).
    \textsuperscript{2} RWTH Aachen University, Germany (e-mail: zlatan.ajanovic@ml.rwth-aachen.de).
    }%
}
\begin{document}
\maketitle
\thispagestyle{empty}
\pagestyle{empty}

\begin{abstract}
    In robot manipulation, Reinforcement Learning (RL) often suffers from low sample efficiency and uncertain convergence, especially in large observation and action spaces.
    Foundation Models (FMs) offer an alternative, demonstrating promise in zero-shot and few-shot settings. However, they can be unreliable due to limited physical and spatial understanding.
    We introduce ExploRLLM, a method that combines the strengths of both paradigms. In our approach, FMs improve RL convergence by generating policy code and efficient representations, while a residual RL agent compensates for the FMs' limited physical understanding.
    We show that ExploRLLM outperforms both policies derived from FMs and RL baselines in table-top manipulation tasks.
    Additionally, real-world experiments show that the policies exhibit promising zero-shot sim-to-real transfer.
    Supplementary material is available at \link{https://explorllm.github.io}.
\end{abstract}

\section{Introduction}

Foundation Models (FMs)~\cite{di2023towards}, which refer to models trained on large-scale data, have shown great potential in robotics.
In particular, language-based FMs, such as Large Language Models (LLMs) and Vision-Language Models (VLMs), are increasingly used in the field.
Large Language Models, such as GPT-4~\cite{openai2023gpt4}, can generate commonsense-aware reasoning in various scenarios. 
For instance, LLMs have demonstrated zero-shot planning capabilities~\cite{huang2022language}, breaking down complex tasks into detailed step-by-step plans without additional training.
When integrated with VLMs, LLMs leverage cross-domain knowledge for robot perception and planning in manipulation tasks~\cite{zeng2022socratic}. 
This synergy allows for extracting environmental affordances and constraints, forming a foundation for subsequent robotic planning~\cite{huang2023voxposer}.
Despite the impressive results of FMs, unpredictable failures in LLM predictions can still lead to robotic errors, and LLMs generally do not learn from past experiences~\cite{carta2023grounding, kambhampati2024llms}.

\begin{figure}
    \centering
    \includegraphics[width=\columnwidth]{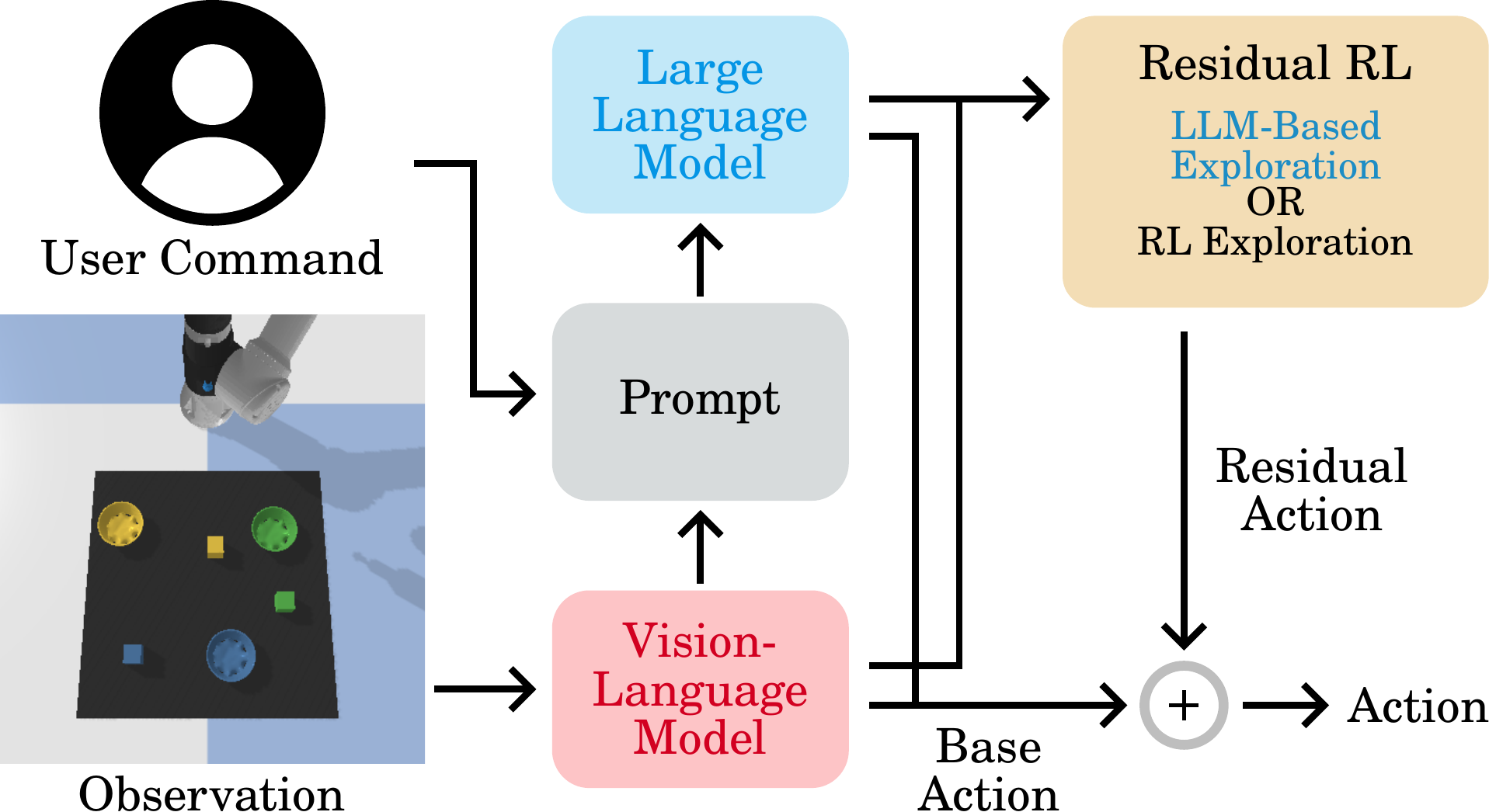} 
    \caption{Graphical overview of ExploRLLM.}
    \label{fig:overview}
\end{figure}

On the other hand, Reinforcement Learning (RL) offers a powerful framework for learning decision-making and control policies through interaction with the environment~\cite{kober2013reinforcement}.
However, RL struggles with the ``curse of dimensionality,'' where large observation and action spaces slow down exploration and convergence. 
To address this, we propose combining FMs and RL by using FMs to guide the RL agent’s exploration as depicted in Figure~\ref{fig:overview}.
While actions generated by FMs may be suboptimal or fail, they can highlight meaningful regions in the action space for exploration.
Traditional RL exploration strategies (e.g., $\epsilon$-greedy, Boltzmann exploration~\cite{sutton2018reinforcement}) are stochastic, focusing on exploration-exploitation trade-offs, but lack mechanisms to incorporate prior knowledge for faster convergence.
Instead, we use LLMs as few-shot planners, generating actions that serve as exploration steps in RL, increasing the likelihood of successful states and gathering more relevant state-action pairs for off-policy RL agents.

Our method, ExploRLLM, improves performance by compensating for FMs' sub-optimality and biases through RL, while FMs accelerate RL training by reducing observation spaces and guiding exploration.
To summarize, our main contributions are the following.
\begin{enumerate}
    \item We propose ExploRLLM, which employs an RL agent with a) residual action and observation spaces based on affordances identified by FMs and b) LLM-guided exploration.
    \item We introduce a prompting method for LLM-based exploration using hierarchical language-model programs, leading to faster convergence.
    \item We show that ExploRLLM outperforms policies derived solely from LLMs and VLMs and generalizes to unseen scenarios, tasks, and real-world settings without additional training.
\end{enumerate}

\begin{figure*}
\centering
\includegraphics[width=\textwidth]{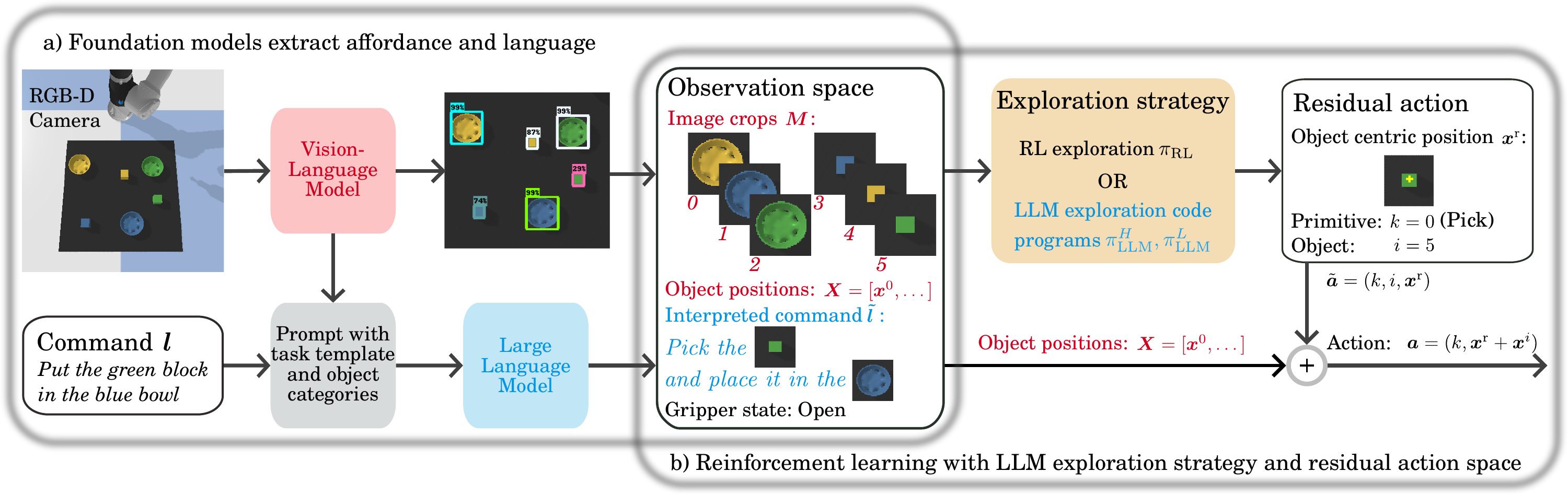} 
\caption{
Implementation structure of ExploRLLM for tabletop manipulation, combining the strengths of RL and FMs.
}
\label{fig:overview_detailed}
\end{figure*}

\section{Related Work}

\subsection{Foundation Models for Planning in Robotics} 

Researchers have shown that LLMs can exhibit reasoning capabilities and generate plans in zero-shot or few-shot settings~\cite{huang2022language,kojima2022large}, which is crucial for high-level planning in robotics. 
These models facilitate task-level planning by integrating environmental groundings, such as affordance value scores~\cite{brohan2023can} or feedback~\cite{huang2022inner}, with their language groundings.
Furthermore, LLMs can generate robot-centric code programs as representations for both task-level~\cite{singh2023progprompt} and skill-level planning~\cite{liang2023code}.
Additionally, VLMs are increasingly integrated into robotics as a perception module of environmental context.
The integration of knowledge from LLMs and VLMs can facilitate the creation of perception-planning pipelines~\cite{zeng2022socratic} and the construction of 3D value maps for zero-shot planning frameworks~\cite{huang2023voxposer}. 
However, due to real-world uncertainty, directly applying VLMs and LLMs to zero-shot tasks may not guarantee success or safety. 
Therefore, in our research, we treat these actions as exploratory behaviors within an RL framework.

\subsection{Foundation Models and Reinforcement Learning}
Incorporating FMs into RL frameworks has notably improved RL's effectiveness.
In~\cite{kwon2023reward}, the authors have implemented LLMs as proxy reward functions, demonstrating their utility in RL. 
In the context of RL for robotics, LLMs are also capable of generating reward signals for robot actions by connecting commonsense reasoning with low-level actions~\cite{yu2023language}, self-refinement~\cite{song2023self} and evolutionary optimization over reward code to enable complex tasks such as dexterous manipulation~\cite{ma2023eureka}. 
Regarding exploration, authors in~\cite{du2023guiding} reward RL agents toward human-meaningful intermediate behaviors by prompting an LLM.
LLMs are also utilized as an intrinsic reward generator to guide exploration for long horizon manipulation tasks~\cite{triantafyllidis2023intrinsic}. 
Contrary to these studies, our approach employs LLM-generated code policies as exploratory actions rather than focusing on reward shaping.
Simultaneously with our study, \cite{chen2024rlingua} introduced a method for improving the sample efficiency of reinforcement learning with LLM-generated rule-based controllers.
In \cite{chen2024rlingua}, the RL policy is regularized towards replay data generated with the LLM policies.
Our method instead uses LLM-generated policies for exploratory actions and does not promote the RL agent to be close to the LLM-generated policies.

\section{Problem Formulation} 

In this study, we focus on language-conditioned tabletop manipulation tasks and a detailed overview of the method is shown in Figure~\ref{fig:overview_detailed}.
Each manipulation task begins at timestep $t=0$ with a linguistically described goal, denoted by $\bm{l}_t$.
The agent receives an observation $\bm{o}_t$, consisting of an overhead RGB-D image and the state of the end-effector. Similar to existing methods (e.g., Transporter~\cite{zeng2021transporter}), the action space involves a pick and a place primitive, denoted as $\{\mathcal{P}_\mathrm{pick}, \mathcal{P}_\mathrm{place}\}$, with each action parameterized by pick and place positions in a top-down view.
We simplify this to a single motion primitive—either pick or place.
This simplification makes the RL problem more tractable by eliminating the need to learn a feature representation for each primitive individually.
The pick or place action is defined as a tuple containing the primitive index $k$ ($0$ for pick, $1$ for place) and a top-down view position, expressed as $\bm{x}$, i.e., $\bm{a}_t = (k_t, \bm{x}_t)$. At each time step, the agent receives a reward $r_t$ consisting of a dense reward component $r^d_t$ and a sparse reward $r^s_t$.

\section{Framework: ExploRLLM}

\subsection{Observation and Action Spaces}

Our method leverages the strengths of LLMs and VLMs to reduce the observation space used for the RL framework.
First of all, the LLM reformulates user-provided language commands into predefined templates and highlights the objects within these templates to form an interpreted command vector~$\tilde{\bm{l}}_t$.
An example is shown in Figure~\ref{fig:overview_detailed}a, where \textit{``Put the green block in the blue bowl''} is interpreted into the template \textit{``Pick the [pick\_object] and place it in the [place\_object]''}.
It is important to note that, within a given task setting, the number and category of objects do not change.
Utilizing VLMs as open-vocabulary object detectors, our system identifies and encloses objects relevant to the task within bounding boxes from the image, represented by their locations $\bm{X}_t = [\bm{x}^0_t, \bm{x}^1_t,... ] $.
RGB-D visual inputs are segmented into crops based on bounding box positions, denoted as $\bm{M}_t = [\bm{m}^0_t, \bm{m}^1_t, ...]$. This method improves the system's robustness to detection-inaccuracies and varying object shapes.
The interpreted commands~$\tilde{\bm{l}}_t$, the positional data~$\bm{X}_t$ and the image patches~$\bm{M}_t$ are then integrated into the reformulated RL observation~$\bm{s}_t$ together with the robot gripper state (open/closed). 
\label{sec: 4a}

As the VLM already extracts each object's position $\bm{x}^i_t$, the action space is converted into an object-centric residual action space (see Figure~\ref{fig:overview_detailed}b). 
The reformulated action space consists of a primitive index $k$, an object index $i$ and a residual position $\bm{x}^\mathrm{r}$, expressed as~$\tilde{\bm{a}}_t = (k_t, i_t, \bm{x}^\mathrm{r}_t)$.
This residual position is then added to the position of the object $i$, i.e., $\bm{x}_t= \bm{x}^i_t + \bm{x}^\mathrm{r}_t$.
This residual action allows the agent to pick or place objects at specific locations. 
This is, for example, needed when picking the letter O, and $\bm{x}^i_t$ denotes the center of the bounding box.
In this case, the residual action $\bm{x}^\mathrm{r}_t$ is needed to prevent picking the letter O at its empty center.
\label{sec: 4b}

\begin{figure*}[t]
    \centering
    \includegraphics[width=0.95\linewidth]{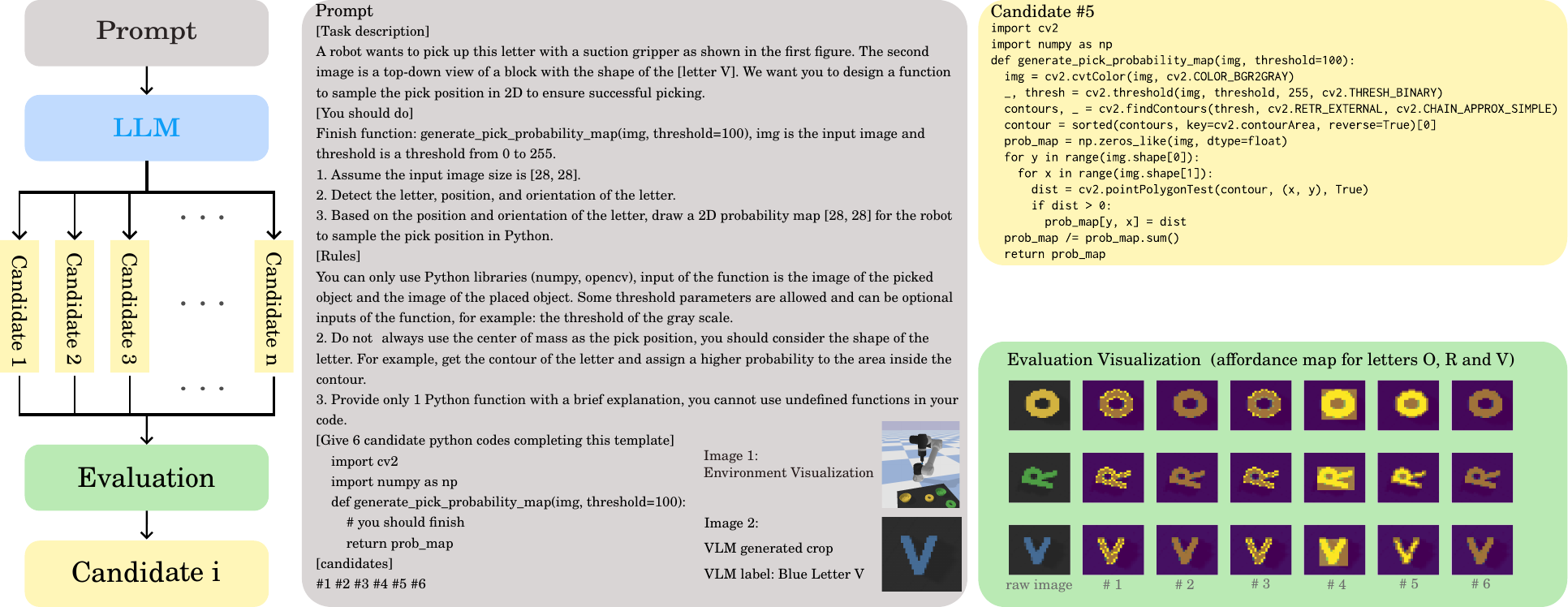}
    \caption{Based on an exploration prompt, candidate policy code is generated. The exploration policy is selected after evaluation.}
    \label{fig:prompt}
\end{figure*}

\begin{algorithm}[t]
    \DontPrintSemicolon
    \caption{Exploration strategy $\pi_\mathrm{EXP}$}
    \label{alg:algorithm1}
    
    \KwIn{state $\bm{s}_t$, high-level LLM policy $\pi_\mathrm{LLM}^H$, low-level LLM policy $\pi_\mathrm{LLM}^L$, RL policy $\pi_\mathrm{RL}$}
    \KwOut{action $\tilde{\bm{a}}_t$}
    \SetKw{KwParam}{Parameter:}
    \KwParam{\textnormal{threshold} $\epsilon$}
    
    $j \sim U_{[0,1)}$ \tcp*{Uniform sampling}
    
    \uIf{$j \leq \epsilon$} {
        $\bm{a}^H_t = (k_t, i_t) \gets \pi_\mathrm{LLM}^H(\bm{s}_t)$ \tcp*{High-level}
        $\bm{x}^\mathrm{r}_t \gets \pi_\mathrm{LLM}^L(\bm{s}_t, \bm{a}^H_t)$         \tcp*{Low-level}
        $\tilde{\bm{a}}_t = (k_t, i_t, \bm{x}^\mathrm{r}_t)$
    }
    \Else {
        $\tilde{\bm{a}}_t \gets \pi_\mathrm{RL}(\bm{s}_t)$ \tcp*{RL policy}
    }
    \Return $\tilde{\bm{a}}_t$
\end{algorithm}

\subsection{LLM-Based Exploration}

Traditional deep RL algorithms (e.g., SAC~\cite{haarnoja2018soft}, PPO~\cite{schulman2017proximal}) do not inherently promote frequent visits to high-value states in high-dimensional state-action spaces, making vision-based tabletop manipulation tasks particularly challenging. In such cases, RL agents may struggle when successful outcomes are rare. Leveraging the planning capabilities of LLMs and the perception strengths of VLMs can help guide the exploration process more effectively by tapping into the rich prior knowledge within these FMs.
The LLM-based exploration strategy, denoted as $\pi_\mathrm{EXP}$ in Algorithm~\ref{alg:algorithm1}, draws inspiration from the $\epsilon$-greedy strategy. 
Specifically, during the rollout collection at each timestep, the off-policy RL agent employs the LLM-based exploration technique if a sampled random variable falls below the threshold $\epsilon$.
Otherwise, the action is selected according to the current RL agent's policy, $\pi_\mathrm{RL}$, as detailed in Algorithm~\ref{alg:algorithm1}.

Inspired by Code-as-Policy (CaP)~\cite{liang2023code}, our method employs the LLM to generate hierarchical language model programs, which are executed during the training phase as exploratory actions.
The hierarchical language model programs include high-level $\pi_\mathrm{LLM}^H$ and low-level $\pi_\mathrm{LLM}^L$ policy code programs. 
A high-level plan primarily involves selecting robot action primitives and the objects to interact with based on the current state of the robot and the objects.

In contrast to high-level tasks, instructing low-level actions poses a more significant challenge because high-level states and actions are more accessible and can be represented as language. 
When dealing with low-level actions, the complexity of the state becomes considerably more intricate, particularly for image-based problems. 
Therefore, instead of a deterministic code policy, we instruct the LLM to produce a code policy $\pi_\mathrm{LLM}^L$ for generating an affordance map according to the input image. 
The low-level exploration behavior is derived from a stochastic policy that relies on the values within this affordance map.
Although the code generated by LLMs lacks guaranteed feasibility and accuracy in robot environments, these models can generate potentially useful policy candidates, with the one exhibiting the highest success rate being selected as shown in Figure~\ref{fig:prompt}.
\label{sec: 4c}
\section{Implementation}

\subsection{RL Agent} 
We use the Soft Actor-Critic (SAC) algorithm with modifications in the collecting rollout phase, detailed in Algorithm~\ref{alg:algorithm1}. 
Other implementation aspects remain consistent with the standard SAC approach in stable-baselines3~\cite{raffin2021stable}.
We employ two convolutional layers to transform every image patch into a vector~$ \bm{\phi}\in \mathbb{R}^{n \times d} $, where n is the number of objects captured by VLM and $d$ the dimension of each patch as encoded by the CNN. 
The vector is subsequently concatenated with the position, robot gripper state, and the extracted episodic language goal $\bm{\tilde{l}}$  to form a new vector  $\bm{\phi}^\prime \in \mathbb{R}^{n \times d^\prime}$, where $d^\prime$ denotes the dimension of each patch's vector following encoding and concatenation.
It then goes to a self-attention layer.
The output features from this layer then go into a two-layer MLP.
The structure mentioned above is consistently utilized across all actor and critic networks.

\subsection{VLM Detection}
Utilizing an open-vocabulary object detector ViLD~\cite{gu2021open}, objects in the environment can be identified by given specific labels. 
However, implementing this model online during training is time-consuming, so ViLD is utilized solely in the evaluation phase. 
In the training phase, the ground truth in the simulation is used to determine the center positions of the bounding boxes. 
It is important to note that ViLD's position detection in real-world scenarios is not always flawless.
To simulate this imperfection, Gaussian noise with a standard deviation equal to half the radius of the image crop is applied to the ground truth positions.

\subsection{LLM Code Policy Generation}
The policy code for executing high-level behavior is obtained using a few-shot prompt in GPT-4~\cite{openai2023gpt4}. 
It includes a list of available robot motion primitives to demonstrate the robot's actions. 
A custom API is also provided to aid the LLM in reasoning, such as determining whether an object is held in the robot's gripper or understanding the relationships between different objects. 
Following the approach demonstrated by~\cite{liang2023code}, where LLMs have been shown capable of generating novel policy codes with example codes and commands, our prompt also includes examples.
They are designed to guide the LLM in formulating plans and conducting geometric reasoning for our specific task scenarios.

For low-level exploration actions, we employ GPT-4 with Vision~\cite{openai2023gpt4}, which generates code using prompts that combine example images with language descriptions, enriching the context with visual information, as shown in Figure~\ref{fig:prompt}. 
The provided example images include a depiction of the environmental setup featuring the robot, a simulated background, objects, and a specific example of image patches inside VLM bounding boxes. 
The prompt describes the requirements and guidelines, enabling generated code to create a probability affordance heatmap for the specified image patch, utilizing external libraries like OpenCV and NumPy.
However, as indicated in Figure~\ref{fig:prompt}, there are instances where the generated affordance map may not be optimal. 
For example, the optimal pick position for the letter O should be at its rim, whereas the heatmap suggests the center. 

To address sub-optimality, we use a stochastic policy based on the affordance map instead of a deterministic one that selects the point of highest affordance. Since RL improves through rewards from environmental interactions, sub-optimal exploration policies can be corrected via learning. This approach also allows for the generation of counter-examples during replay buffer collection.

\begin{table*}[t]
\centering
\caption{
Results of 50 evaluation episodes for short-horizon (SH), long-horizon (LH), and different initialization methods: no object overlap (NO) and allowed overlap (AO). ExploRLLM standard deviations are shown for 6 seeds.
}
\label{table1}
\begin{tabular}{c|c|c|c|c|c|c|c|c}
\hline
\multirow{2}{*}{Method} & \multicolumn{4}{c|}{Overall success rate }&\multicolumn{4}{c}{Low-level error rate }\\
\cline{2-9}

&  SH NO & SH AO & LH NO & LH AO &  SH NO & SH AO & LH NO & LH AO \\
\hline
ExploRLLM (20\%) & 0.86$\pm$0.05 & 0.80$\pm$0.06 &	0.70$\pm$0.11 &	0.54$\pm$0.09 &	0.14$\pm$0.05 &	0.20$\pm$0.06 &	0.18$\pm$0.10 &	0.22$\pm$0.9 \\

ExploRLLM (0\%) & 0.56$\pm$0.40 & 0.48$\pm$0.36 &	-- & -- &	0.32$\pm$0.24 & 0.42$\pm$0.30 & -- & -- \\

CaP$^*$ & 0.60 & 0.48 &	0.38 & 0.30 & 0.38 &	0.52&	0.42 &	0.48  \\

Socratic Models + CLIPort & 0.78 & 0.64 &	0.50 & 0.36 &	0.22 &	0.28&	0.22 &  0.28  \\

Inner Monologue + CLIPort & 0.82 & 0.72 &	0.58 & 0.42 &	0.18 &	0.26&	0.20 &	0.24  \\

\hline
\end{tabular}
\label{table_com}
\end{table*}

\begin{figure*}[!ht]
     \centering
     \begin{subfigure}[b]{0.49\textwidth}
         \centering
         \includegraphics[width=\textwidth]{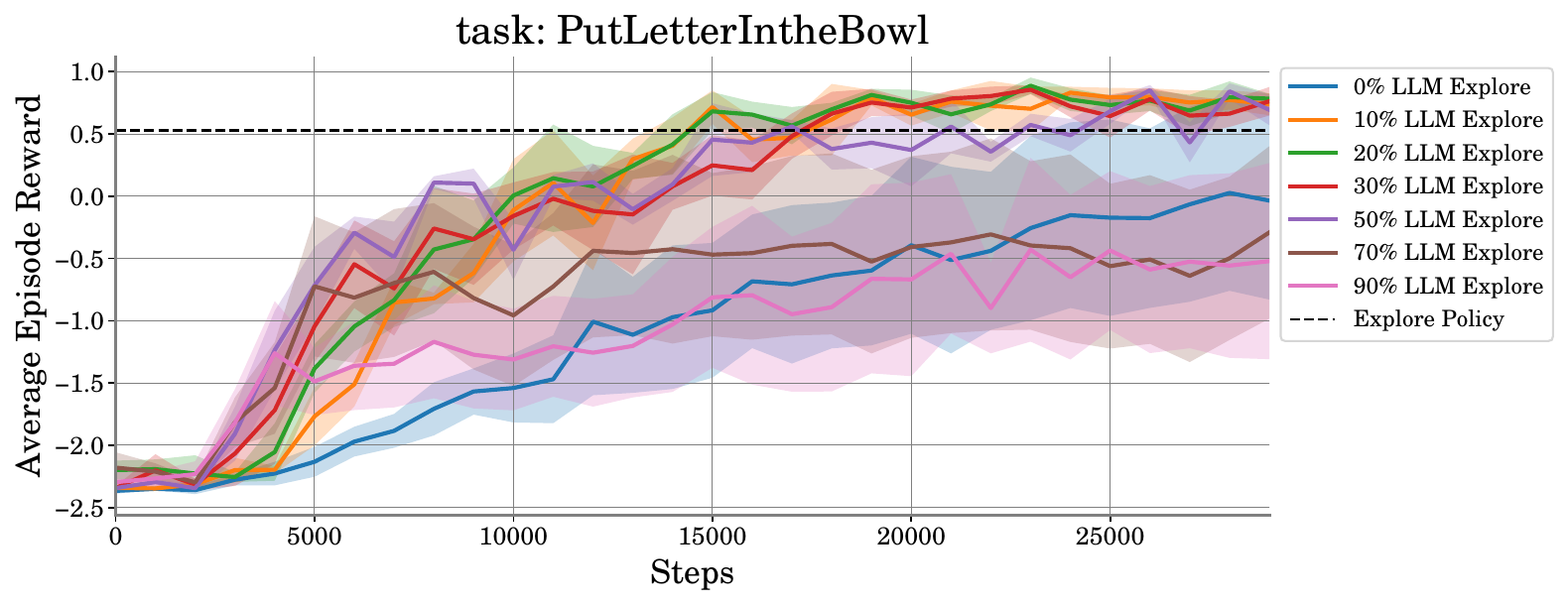}
         \caption{\textit{Pick the [pick letter]  and place it in the [place color] bowl} (SH).}
         \label{fig:train short-horizon task}
     \end{subfigure}
     \hfill
     \begin{subfigure}[b]{0.49\textwidth}
         \centering
         \includegraphics[width=\textwidth]{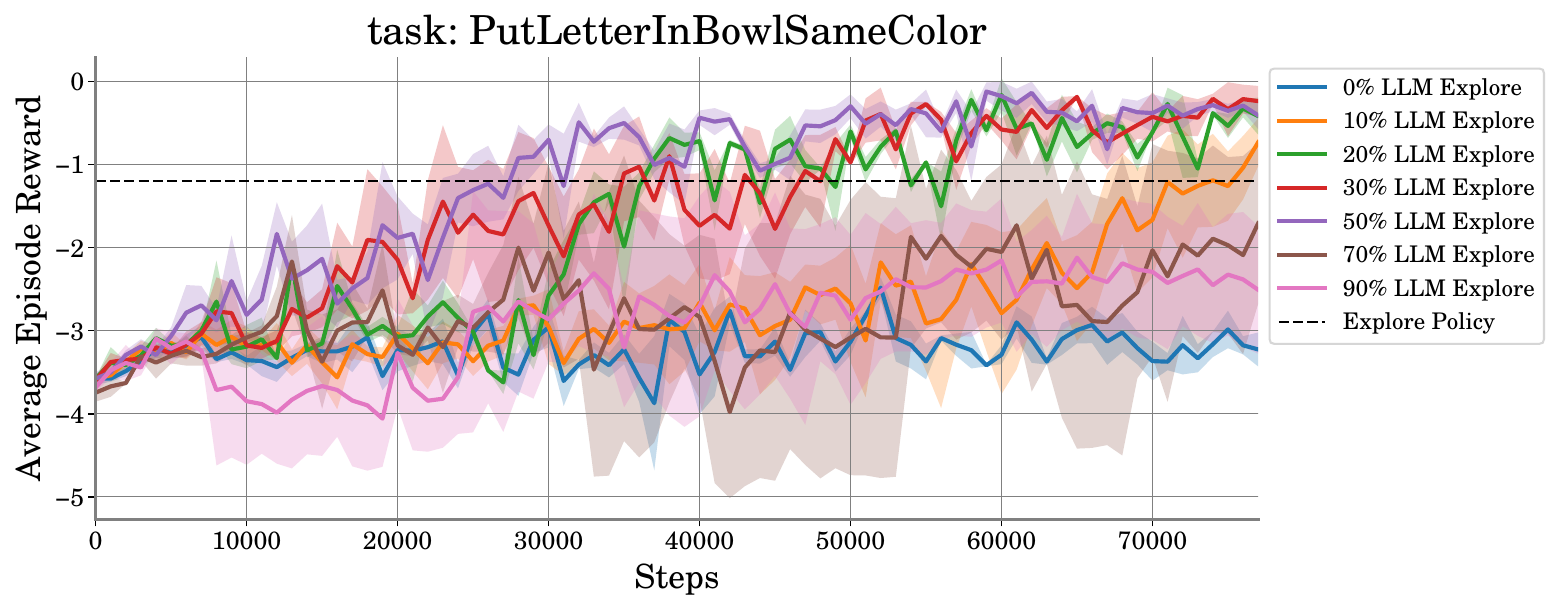}
         \caption{\textit{Put all letters in the bowl of the corresponding color} (LH).}
         \label{fig:train Long-horizon task}
     \end{subfigure}
     \caption{Training curves for varying exploration rates in SH and LH tasks.
     ExploRLLM outperforms the exploration policies (dashed lines) and RL without LLM-based exploration ($\epsilon=0$).
     In the LH task, LLM-based exploration is crucial for success.}
     \label{fig6}
\end{figure*}

\section{Experimental Setups}

\subsection{Simulation Setup}

We evaluated the proposed method on a simulated tabletop pick-and-place task, as shown in Figure~\ref{fig:overview_detailed}. 
Similar to \cite{zeng2021transporter} and \cite{shridhar2022cliport}, we use a UR5e, and the input observation is a top-down RGB-D image.
Inspired by \cite{shridhar2022cliport}, we increased the task difficulty by replacing simple blocks with various objects, such as letters. We assess our method in two tasks: a short-horizon (SH) task, \textit{``Pick the [pick\_letter] and place it in the [place\_color] bowl''}, and a long-horizon (LH) task, \textit{``Put all letters in the bowl of the corresponding color''}, as shown in Figure ~\ref{fig: robot}.
In the SH task, each episode starts with three letters and three bowls randomly placed on the table, with pick-and-place actions generated from random language commands.
The task is completed when the robot places the chosen letter in the specified bowl.
In the LH task, all letters and bowls are randomly arranged, and the task is completed when each letter is placed in a bowl that matches its color.

\subsection{Real-World Setup}
We validated our approach on a Franka Panda robot equipped with a suction gripper and an RGB-D camera, as shown in Figure~\ref{fig: robot}, implementing our policy and code in the EAGERx~\cite{vanderheijden2024eagerx} framework.
Given the potential risks to hardware and the time-intensive nature of direct training,  we completed training in simulation, with real-robot applications limited to evaluation.
We used ViLD to identify bounding boxes based on object names.  
To simulate real-world conditions more accurately, we introduced noise to the bounding box center's position during the training phase in the simulation, mimicking the positional uncertainty inherent in VLM detection. 
We also added noise to bounding box positions and image inputs, simulating VLM detection uncertainty and camera noise, including lighting variations.

\section{Results}

\begin{table}[t]
  \begin{center}
    \caption{ExploRLLM training returns for varying $\epsilon$.}
    \resizebox{\columnwidth}{!}{%
        \begin{tabular}{c c c} 
        \hline
         Explore $\epsilon\ (\%)$ & SH Task (25k steps)& LH Task (75k steps)\\
         \hline
         $0$  & $ - 0.03 \pm 1.13$ & $ - 3.22 \pm 0.29$\\
         $10$ & $   0.74 \pm 0.13$ & $ - 0.73 \pm 0.40$\\
         $20$ & $   0.79 \pm 0.06$ & $ - 0.42 \pm 0.31$\\
         $30$ & $   0.76 \pm 0.16$ & $ - 0.23 \pm 0.26$\\
         $50$ & $   0.70 \pm 0.17$ & $ - 0.40 \pm 0.23$\\
         $70$ & $ - 0.29 \pm 0.98$ & $ - 1.71 \pm 1.38$\\
         $90$ & $ - 0.52 \pm 1.12$ & $ - 2.51 \pm 1.09$\\
         \hline
         Exploration Policy & 0.53 & -1.2\\
        \hline
        \end{tabular}
        }
        \label{training table}
  \end{center}
\end{table}

\begin{table}[t]
  \begin{center}
    \caption{Success rate (\%) of SH ExploRLLM with \cite{zeng2022socratic}.}
    \resizebox{\columnwidth}{!}{%
        \begin{tabular}{c c c c} 
        \hline
         Task Settings & Seen & Unseen Color & Unseen Letters \\
         \hline
          Socratic Models + ExploRLLM & 74 & 68 & 56\\
          Socratic Models + CLIPort & 72 & 50 & 34\\
         
        \hline
        \end{tabular}
        }
        \label{sm table}
  \end{center}
\end{table}

\begin{figure}[t]
\centering
\includegraphics[width=0.9\linewidth]{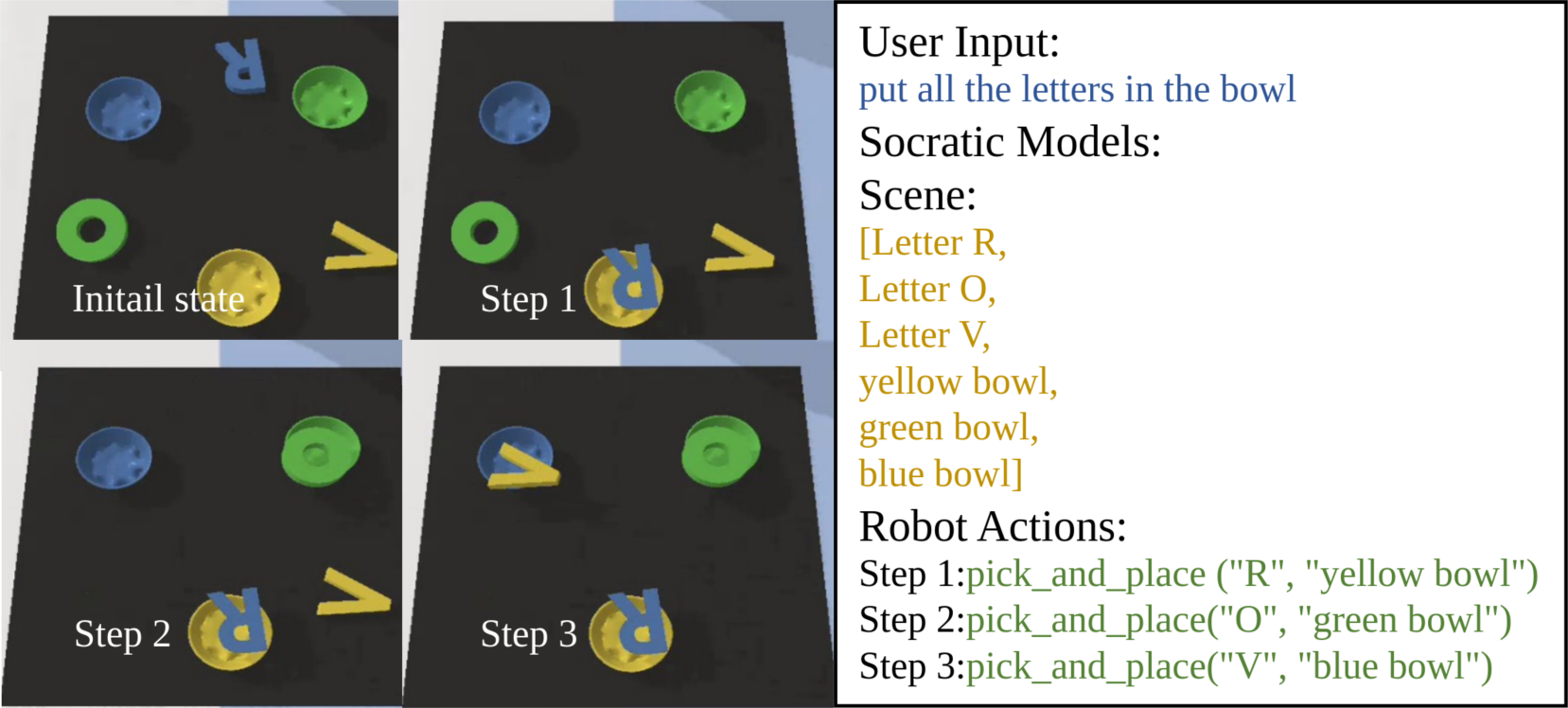}
\caption{Short-horizon ExploRLLM policies can be used in long-horizon tasks with zero-shot LLM planners, e.g., \cite{zeng2022socratic}.}
\label{fig7}
\end{figure}

\begin{figure*}[t]
     \centering
     \hfill
     \begin{subfigure}[b]{0.28\textwidth}
         \centering
         \includegraphics[width=\textwidth]{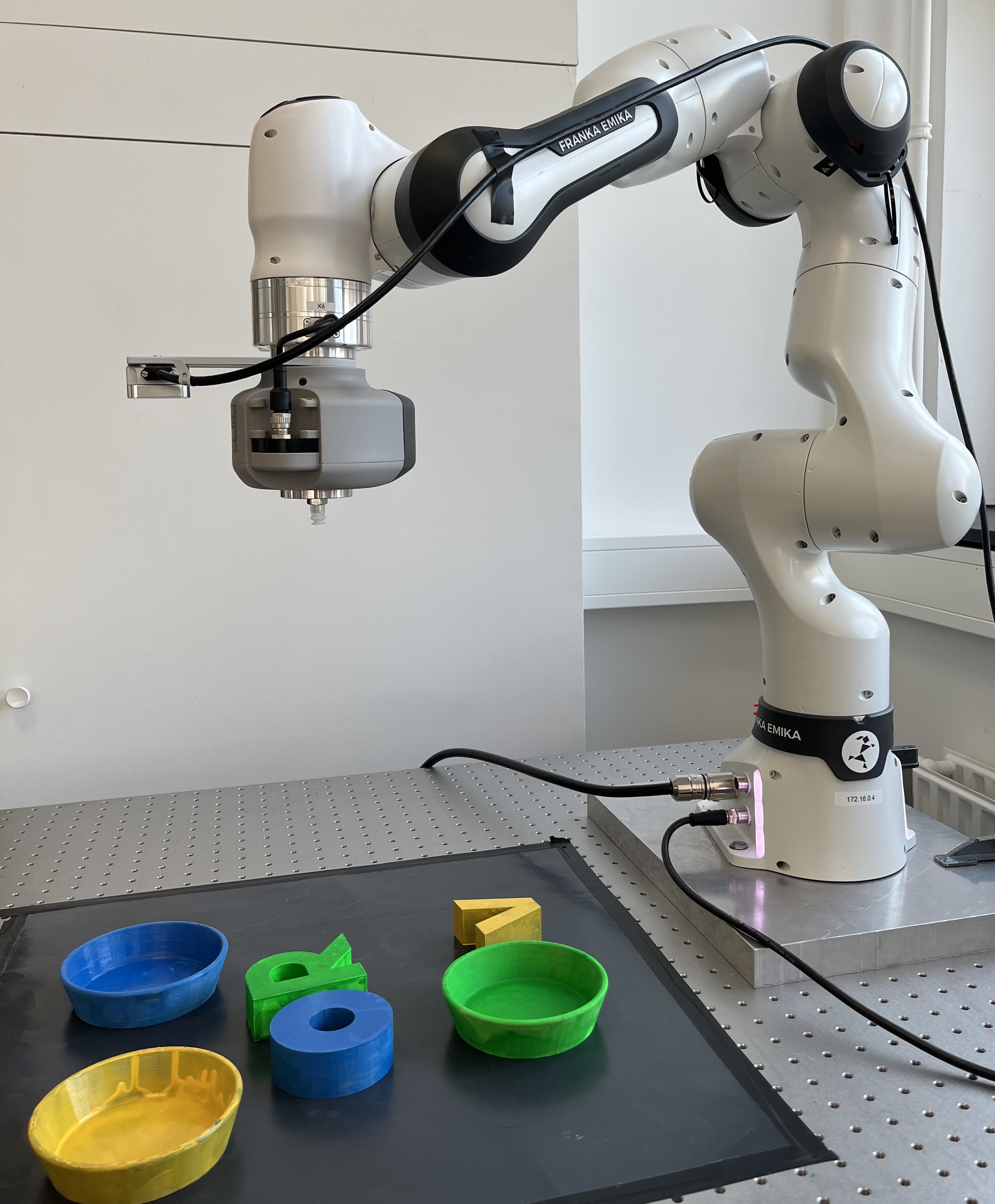}
         \caption{Real-world experimental setup.}
         \label{fig: robot}
     \end{subfigure}
     \hfill
     \begin{subfigure}[b]{0.71\textwidth}
         \centering
         \includegraphics[width=\textwidth]{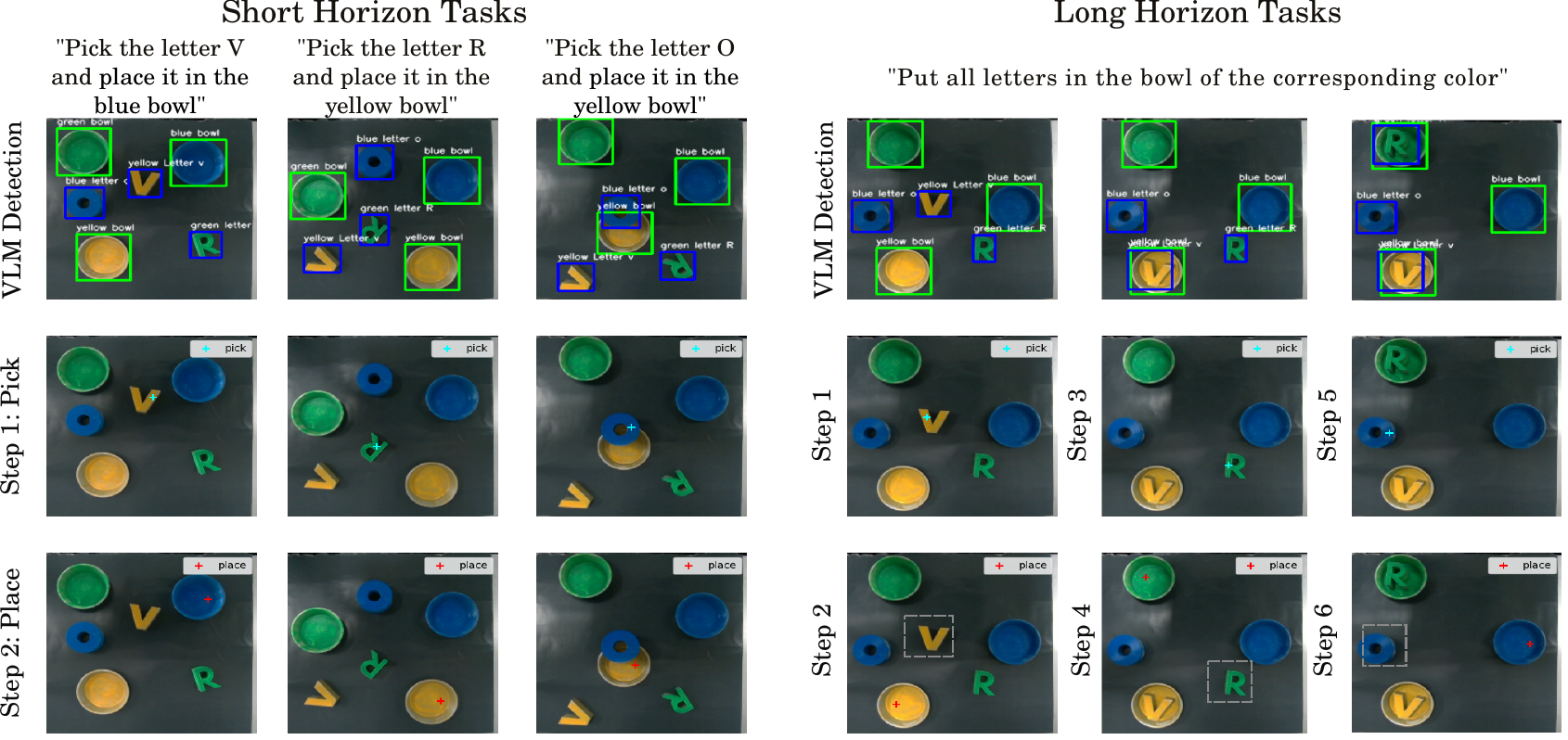}
         \caption{Visualization of VLM detections and pick and place actions.}
         \label{fig: pick place pos}
     \end{subfigure}
     \caption{ExploRLLM can be practically applied using a sim-to-real approach with transfer due to VLM object detections.}
\end{figure*}

\subsection{Simulation Results}

We investigated the effect of varying LLM-based exploration frequencies on training convergence, using $\epsilon \in \{0.0, 0.1, \dots, 0.9\}$, as shown in Figure~\ref{fig6}. An $\epsilon$ of 0 corresponds to standard SAC.
We trained the agents with six random seeds per frequency, and each session began with a 20,000-step warm-up phase without LLM exploration, as no significant policy improvements were observed during this phase. Post-warm-up results, shown in Figure~\ref{fig6} and detailed in Table~\ref{training table} for both short- and long-horizon tasks, indicate that ExploRLLM consistently outperforms LLM-only policies across various exploration frequencies.
\label{sec: 6a}

In the short-horizon task (Figure~\ref{fig:train short-horizon task}), training without LLM-based exploration is often unstable, resulting in either a successful policy or failure to converge within the duration of our experiments.
Training stabilizes and converges faster when the exploration frequency is within $0 < \epsilon \leq 0.5$, with minimal variation across different $\epsilon$ values.
However, increasing $\epsilon$ beyond 0.5 reduces the proportion of online data, slowing progress and introducing greater instability into the training.
For long-horizon tasks, Figure~\ref{fig:train Long-horizon task} shows that higher frequencies of LLM-based exploration ($0 < \epsilon \leq 0.5$) correlate with faster training.
These results highlight the importance of LLM-based exploration in navigating complex tasks by guiding experience toward the optimal region, thereby mitigating challenges from large observation and action spaces.
However, similar to the short-horizon tasks, excessive exploration rates introduce instability and fail to converge within the duration of the experiments.

To evaluate the effectiveness of ExploRLLM, we benchmark its performance against four baselines: ExploRLLM without the LLM-based exploration policy, the CaP-style policy~\cite{liang2023code} (our exploration policy), Socratic Models~\cite{zeng2022socratic}, and Inner Monologue~\cite{huang2022inner}.
Our Socratic Models and Inner Monologue implementations use ViLD~\cite{gu2021open} as the object detector and GPT-4~\cite{openai2023gpt4} as a multi-step planner. The individual steps are executed by a pre-trained CLIPort~\cite{shridhar2022cliport} model with 500 demonstrations.
The key difference between Socratic Models and Inner Monologue is that Inner Monologue features a success detector that can identify mistakes.

During evaluation, the letter colors range from seen to unseen colors. Tasks and initialization methods vary, with “NO” indicating no overlap between the initial positions of letters and bowls and “AO” allowing overlaps. These configurations assess each method's robustness in handling complex object relationships.

For short-horizon tasks, as shown in Table~\ref{table_com}, ExploRLLM maintains stable performance.
In contrast, versions without the exploration policy have not all converged and exhibit high variance in success rates and low-level errors. Our method surpasses other methods for LLM-generated policies in success rates, reduces robot behavior errors, and minimizes the performance gap between NO and AO scenarios, emphasizing the exploration policy's role in correcting FMs' inaccuracies. In contrast, CLIPort-based methods struggle with novel scenarios or complex geometric object relationships.
For long-horizon tasks, RL agents without LLM-based exploration fail to converge within the duration of the experiment.
As shown in Table~\ref{table_com}, ExploRLLM outperforms Socratic Models, Inner Monologue, and LLM-generated policies, achieving superior results in long-horizon tasks.
\label{sec: 6b}

Although our short-horizon agent is trained specifically for a pre-defined pick-and-place task, our approach can transfer to unseen long-horizon tasks in similar environments. This is made possible by integrating a zero-shot planner framework, such as Socratic Models~\cite{zeng2022socratic}.
This framework effectively breaks down user-provided input into individual action steps, each serving as a distinct language command for our single-step RL agent, as illustrated in Figure \ref{fig7}. 
Following the execution of each command, the task space is reset, allowing for the subsequent command to be executed.
Apart from unseen colors, unseen letters are also included to evaluate the generalization capabilities of unseen scenarios.
Table~\ref{sm table} demonstrates that the short-horizon ExploRLLM adapts to these settings, surpassing earlier Socratic Models versions.
Using VLMs to provide bounding boxes and positions, our approach reformulates the observation space, enabling RL to focus on learning the physical attributes of objects, which is crucial for precise pick-and-place tasks. This strategy minimizes distractions from variations in colors and shapes.
\label{sec: 6c}

\subsection{Real-World Results}

We evaluated ExploRLLM in two real-world scenarios: one replicating all letters from the simulation and another introducing the unseen letter `C'.
Each scenario was tested over 15 episodes.
The short-horizon ExploRLLM achieved success rates of 66.6\% for seen letters and 53.3\% for the unseen letter scenario. In comparison, the long-horizon ExploRLLM recorded success rates of 40\% for seen letters and 33.3\% for unseen letters. Despite the sim-to-real gap, our approach shows promising results without additional real-world training. As the VLM extracts the observation space, the RL agent trained in simulation is less distracted by real-world noise. Figure~\ref{fig: pick place pos} illustrates the adaptability of our method in handling diverse object orientations, understanding logical relationships between objects, and executing long-horizon tasks in real-world settings. However, challenges remain with noise in the color and depth perception of objects, which hampers the RL agent's ability to manipulate objects. Using a photorealistic simulator with extensive domain randomization is expected to improve performance.
\label{sec 6d}

\section{Conclusion and Discussion}

In this work, we presented ExploRLLM, a method that combines RL with FMs.
ExploRLLM accelerates RL convergence by using actions informed by LLMs and VLMs to guide exploration, demonstrating the benefits of integrating the strengths of both RL and FMs. We evaluated our method on tabletop manipulation tasks, showing superior success rates compared to policies based solely on LLMs and VLMs. ExploRLLM also generalizes unseen colors, letters, and tasks better.
Ablation experiments with varying levels of LLM-guided exploration indicated that extensive tuning of this parameter is unnecessary as values of $0 < \epsilon \le 0.5$ showed convergence improvements.
Additionally, we validated the method's ability to transfer learned policies from simulation to real-world scenarios without additional training through real robot experiments.
Currently, our framework focuses on tabletop manipulation, but we plan to extend it to a broader range of robotic manipulation tasks. While the system can correct low-level robotic actions, it struggles with mitigating high-level errors that are less frequent in simulations. Future work will focus on addressing these high-level discrepancies.

\section{Acknowledgments}

Research reported in this work was partially or completely facilitated by computational resources and support of the Delft AI Cluster (DAIC) \cite{DAIC} at TU Delft (RRID: SCR\_025091), but remains the sole responsibility of the authors, not the DAIC team.


\bibliographystyle{IEEEtran}
\bibliography{bib}

\begin{thebibliography}{10}
\providecommand{\url}[1]{#1}
\csname url@samestyle\endcsname
\providecommand{\newblock}{\relax}
\providecommand{\bibinfo}[2]{#2}
\providecommand{\BIBentrySTDinterwordspacing}{\spaceskip=0pt\relax}
\providecommand{\BIBentryALTinterwordstretchfactor}{4}
\providecommand{\BIBentryALTinterwordspacing}{\spaceskip=\fontdimen2\font plus
\BIBentryALTinterwordstretchfactor\fontdimen3\font minus
  \fontdimen4\font\relax}
\providecommand{\BIBforeignlanguage}[2]{{%
\expandafter\ifx\csname l@#1\endcsname\relax
\typeout{** WARNING: IEEEtran.bst: No hyphenation pattern has been}%
\typeout{** loaded for the language `#1'. Using the pattern for}%
\typeout{** the default language instead.}%
\else
\language=\csname l@#1\endcsname
\fi
#2}}
\providecommand{\BIBdecl}{\relax}
\BIBdecl

\bibitem{di2023towards}
N.~Di~Palo, A.~Byravan, L.~Hasenclever, M.~Wulfmeier, N.~Heess, and
  M.~Riedmiller, ``Towards a unified agent with foundation models,'' in
  \emph{Workshop on Reincarnating Reinforcement Learning at ICLR}, 2023.

\bibitem{openai2023gpt4}
OpenAI, ``{GPT-4} technical report,'' 2023.

\bibitem{huang2022language}
W.~Huang, P.~Abbeel, D.~Pathak, and I.~Mordatch, ``Language models as zero-shot
  planners: Extracting actionable knowledge for embodied agents,'' in
  \emph{International Conference on Machine Learning (ICML)}.\hskip 1em plus
  0.5em minus 0.4em\relax PMLR, 2022.

\bibitem{zeng2022socratic}
A.~Zeng, M.~Attarian, B.~Ichter, K.~M. Choromanski, A.~Wong, S.~Welker,
  F.~Tombari, A.~Purohit, M.~S. Ryoo, V.~Sindhwani, J.~Lee, V.~Vanhoucke, and
  P.~Florence, ``Socratic models: Composing zero-shot multimodal reasoning with
  language,'' in \emph{International Conference on Learning Representations
  (ICLR)}, 2023.

\bibitem{huang2023voxposer}
W.~Huang, C.~Wang, R.~Zhang, Y.~Li, J.~Wu, and L.~Fei-Fei, ``{VoxPoser}:
  Composable {3D} value maps for robotic manipulation with language models,''
  in \emph{Conference on Robot Learning (CoRL)}.\hskip 1em plus 0.5em minus
  0.4em\relax PMLR, 2023.

\bibitem{carta2023grounding}
T.~Carta, C.~Romac, T.~Wolf, S.~Lamprier, O.~Sigaud, and P.-Y. Oudeyer,
  ``Grounding large language models in interactive environments with online
  reinforcement learning,'' in \emph{International Conference on Machine
  Learning (ICML)}.\hskip 1em plus 0.5em minus 0.4em\relax PMLR, 2023.

\bibitem{kambhampati2024llms}
S.~Kambhampati, K.~Valmeekam, L.~Guan, M.~Verma, K.~Stechly, S.~Bhambri, L.~P.
  Saldyt, and A.~B. Murthy, ``Position: {LLM}s can{\textquoteright}t plan, but
  can help planning in {LLM}-modulo frameworks,'' in \emph{International
  Conference on Machine Learning (ICML)}, 2024.

\bibitem{kober2013reinforcement}
J.~Kober, J.~A. Bagnell, and J.~Peters, ``Reinforcement learning in robotics: A
  survey,'' \emph{The International Journal of Robotics Research}, vol.~32,
  no.~11, pp. 1238--1274, 2013.

\bibitem{sutton2018reinforcement}
R.~S. Sutton and A.~G. Barto, \emph{Reinforcement learning: An
  introduction}.\hskip 1em plus 0.5em minus 0.4em\relax MIT press, 2018.

\bibitem{kojima2022large}
T.~Kojima, S.~S. Gu, M.~Reid, Y.~Matsuo, and Y.~Iwasawa, ``Large language
  models are zero-shot reasoners,'' \emph{Advances in Neural Information
  Processing Systems (NeurIPS)}, 2022.

\bibitem{brohan2023can}
A.~Brohan, Y.~Chebotar, C.~Finn, K.~Hausman, A.~Herzog, D.~Ho, J.~Ibarz,
  A.~Irpan, E.~Jang, R.~Julian \emph{et~al.}, ``{Do as I can, not as I say:
  Grounding language in robotic affordances},'' in \emph{Conference on Robot
  Learning (CoRL)}.\hskip 1em plus 0.5em minus 0.4em\relax PMLR, 2023.

\bibitem{huang2022inner}
W.~Huang, F.~Xia, T.~Xiao, H.~Chan, J.~Liang, P.~Florence, A.~Zeng, J.~Tompson,
  I.~Mordatch, Y.~Chebotar, P.~Sermanet, T.~Jackson, N.~Brown, L.~Luu,
  S.~Levine, K.~Hausman, and B.~Ichter, ``Inner monologue: Embodied reasoning
  through planning with language models,'' in \emph{Conference on Robot
  Learning (CoRL)}.\hskip 1em plus 0.5em minus 0.4em\relax PMLR, 2023.

\bibitem{singh2023progprompt}
I.~Singh, V.~Blukis, A.~Mousavian, A.~Goyal, D.~Xu, J.~Tremblay, D.~Fox,
  J.~Thomason, and A.~Garg, ``{ProgPrompt: Generating situated robot task plans
  using large language models},'' in \emph{IEEE International Conference on
  Robotics and Automation (ICRA)}, 2023.

\bibitem{liang2023code}
J.~Liang, W.~Huang, F.~Xia, P.~Xu, K.~Hausman, B.~Ichter, P.~Florence, and
  A.~Zeng, ``Code as policies: Language model programs for embodied control,''
  in \emph{IEEE International Conference on Robotics and Automation (ICRA)},
  2023.

\bibitem{kwon2023reward}
M.~Kwon, S.~M. Xie, K.~Bullard, and D.~Sadigh, ``Reward design with language
  models,'' in \emph{International Conference on Learning Representations
  (ICLR)}, 2023.

\bibitem{yu2023language}
W.~Yu, N.~Gileadi, C.~Fu, S.~Kirmani, K.-H. Lee, M.~G. Arenas, H.-T.~L. Chiang,
  T.~Erez, L.~Hasenclever, J.~Humplik, B.~Ichter, T.~Xiao, P.~Xu, A.~Zeng,
  T.~Zhang, N.~Heess, D.~Sadigh, J.~Tan, Y.~Tassa, and F.~Xia, ``Language to
  rewards for robotic skill synthesis,'' in \emph{Conference on Robot Learning
  (CoRL)}.\hskip 1em plus 0.5em minus 0.4em\relax PMLR, 2023.

\bibitem{song2023self}
J.~Song, Z.~Zhou, J.~Liu, C.~Fang, Z.~Shu, and L.~Ma, ``Self-refined large
  language model as automated reward function designer for deep reinforcement
  learning in robotics,'' \emph{arXiv preprint arXiv:2309.06687}, 2023.

\bibitem{ma2023eureka}
Y.~J. Ma, W.~Liang, G.~Wang, D.-A. Huang, O.~Bastani, D.~Jayaraman, Y.~Zhu,
  L.~Fan, and A.~Anandkumar, ``Eureka: Human-level reward design via coding
  large language models,'' in \emph{International Conference on Learning
  Representations (ICLR)}, 2024.

\bibitem{du2023guiding}
Y.~Du, O.~Watkins, Z.~Wang, C.~Colas, T.~Darrell, P.~Abbeel, A.~Gupta, and
  J.~Andreas, ``Guiding pretraining in reinforcement learning with large
  language models,'' in \emph{International Conference on Machine Learning
  (ICML)}.\hskip 1em plus 0.5em minus 0.4em\relax PMLR, 2023.

\bibitem{triantafyllidis2023intrinsic}
E.~Triantafyllidis, F.~Christianos, and Z.~Li, ``Intrinsic language-guided
  exploration for complex long-horizon robotic manipulation tasks,'' in
  \emph{IEEE International Conference on Robotics and Automation (ICRA)}, 2024.

\bibitem{chen2024rlingua}
L.~Chen, Y.~Lei, S.~Jin, Y.~Zhang, and L.~Zhang, ``Rlingua: Improving
  reinforcement learning sample efficiency in robotic manipulations with large
  language models,'' \emph{IEEE Robotics and Automation Letters}, vol.~9,
  no.~7, pp. 6075--6082, 2024.

\bibitem{zeng2021transporter}
A.~Zeng, P.~Florence, J.~Tompson, S.~Welker, J.~Chien, M.~Attarian,
  T.~Armstrong, I.~Krasin, D.~Duong, V.~Sindhwani \emph{et~al.}, ``Transporter
  networks: Rearranging the visual world for robotic manipulation,'' in
  \emph{Conference on Robot Learning (CoRL)}.\hskip 1em plus 0.5em minus
  0.4em\relax PMLR, 2021.

\bibitem{haarnoja2018soft}
T.~Haarnoja, A.~Zhou, P.~Abbeel, and S.~Levine, ``Soft actor-critic: Off-policy
  maximum entropy deep reinforcement learning with a stochastic actor,'' in
  \emph{International Conference on Machine Learning (ICML)}.\hskip 1em plus
  0.5em minus 0.4em\relax PMLR, 2018.

\bibitem{schulman2017proximal}
J.~Schulman, F.~Wolski, P.~Dhariwal, A.~Radford, and O.~Klimov, ``Proximal
  policy optimization algorithms,'' \emph{arXiv preprint arXiv:1707.06347},
  2017.

\bibitem{raffin2021stable}
A.~Raffin, A.~Hill, A.~Gleave, A.~Kanervisto, M.~Ernestus, and N.~Dormann,
  ``Stable-baselines3: Reliable reinforcement learning implementations,''
  \emph{The Journal of Machine Learning Research}, vol.~22, no.~1, pp.
  12\,348--12\,355, 2021.

\bibitem{gu2021open}
X.~Gu, T.-Y. Lin, W.~Kuo, and Y.~Cui, ``Open-vocabulary object detection via
  vision and language knowledge distillation,'' in \emph{International
  Conference on Learning Representations (ICLR)}, 2022.

\bibitem{shridhar2022cliport}
M.~Shridhar, L.~Manuelli, and D.~Fox, ``{CLIPort: What and where pathways for
  robotic manipulation},'' in \emph{Conference on Robot Learning (CoRL)}.\hskip
  1em plus 0.5em minus 0.4em\relax PMLR, 2022.

\bibitem{vanderheijden2024eagerx}
B.~van~der Heijden, J.~Luijkx, L.~Ferranti, J.~Kober, and R.~Babuska, ``Engine
  agnostic graph environments for robotics ({EAGERx}): A graph-based framework
  for sim2real robot learning,'' \emph{IEEE Robotics and Automation Magazine},
  pp. 2--15, 2024.

\bibitem{DAIC}
\BIBentryALTinterwordspacing
{Delft AI Cluster (DAIC)}, ``{The Delft AI Cluster (DAIC), RRID:SCR\_025091},''
  2024. [Online]. Available: \url{https://doc.daic.tudelft.nl/}
\BIBentrySTDinterwordspacing

\end{thebibliography}

\end{document}